\documentclass[10pt,twocolumn,letterpaper]{article}

\usepackage{wacv}              

\usepackage{graphicx}
\usepackage{amsmath}
\usepackage{amssymb}
\usepackage{booktabs}
\usepackage{graphicx}
\usepackage{wrapfig}
\usepackage{multirow}
\usepackage{animate}
\usepackage{cuted}
\usepackage{booktabs}
\usepackage{appendix}
\usepackage{pifont}  

\usepackage[pagebackref,breaklinks,colorlinks,citecolor=blue]{hyperref}

\usepackage[capitalize]{cleveref}
\crefname{section}{Sec.}{Secs.}
\Crefname{section}{Section}{Sections}
\Crefname{table}{Table}{Tables}
\crefname{table}{Tab.}{Tabs.}

\def\wacvPaperID{1128} 
\def\confName{WACV}
\def\confYear{2026}

\begin{document}

\title{VFace: A Training-Free Approach for Diffusion-Based Video
Face Swapping}


\author{
Sanoojan Baliah$^{1,2}$ \quad
Yohan Abeysinghe$^{1}$ \quad
Rusiru Thushara$^{1}$ \quad
Khan Muhammad$^{3}$ \quad
Abhinav Dhall$^{4}$\\
Karthik Nandakumar$^{1,2}$ \quad
Muhammad Haris Khan$^{1}$\\
$^1$MBZUAI, UAE \quad
$^2$Michigan State University, USA \quad
$^3$Sungkyunkwan University, South Korea\\
$^4$Monash University, Australia\\
{\tt\small 
\{baliahsa,nandakum\}@msu.edu \quad \{yohan.abeysinghe,rusiru.thushara,muhammad.haris\}@mbzuai.ac.ae} \\
{\tt\small khan.muhammad@ieee.org \quad
abhinav.dhall@monash.edu}
}


\maketitle
\begin{strip}
\centering
\begin{minipage}[b]{0.48\textwidth}
    \centering
    \animategraphics[width=0.98\linewidth,height=0.59\textwidth,loop,poster=first]{10}{Videos/teaser1/combined_}{00}{23}
\end{minipage}
\begin{minipage}[b]{0.48\textwidth}
    \centering
\animategraphics[width=0.98\linewidth,height=0.59\textwidth,loop,poster=first]{10}{Videos/elon.jpegcombined/combined_}{00}{23}
\end{minipage}
\vspace{-1em}
\captionof{figure}{Our method (VFace) for video face swapping effectively guides the structure of the target video (T) while preserving the identity features from the source image (S). Video animation can be viewed using Acrobat Reader (Click to play).}
\label{fig:teaser}
\end{strip}

\begin{abstract}
We present a training-free, plug-and-play method, namely VFace, for high-quality face swapping in videos. It
can be seamlessly integrated with image-based face swapping approaches built on diffusion models. First, we introduce a Frequency Spectrum Attention Interpolation technique to facilitate generation and intact key identity characteristics. Second, we achieve Target Structure Guidance via plug-and-play attention injection to better align the structural features from the target frame to the generation.
Third, we present a Flow-Guided Attention Temporal Smoothening mechanism that enforces spatiotemporal coherence without modifying the underlying diffusion model to reduce temporal inconsistencies typically encountered in frame-wise generation.
Our method requires no additional training or video-specific fine-tuning. Extensive experiments show that our method significantly enhances temporal consistency and visual fidelity, offering a practical and modular solution for video-based face swapping. Our code is available at \href{https://github.com/Sanoojan/VFace}{VFace}.




\end{abstract}


\section{Introduction}

Face swapping refers to synthesizing an image or video where the identity of a source face is seamlessly transferred onto a target, while preserving the target’s pose, expression, lighting, and background~\cite{baliah2025reface, wang2025dynamicface}. This capability has gained considerable attention due to its applications in visual effects, digital avatars, privacy preservation, and entertainment~\cite{moussa2025face}. Although substantial progress has been made in static image face swapping, achieving high-fidelity, identity-preserving face swaps in video remains a significant challenge due to issues like temporal flickering, identity drift, and inconsistent expressions ~\cite{wang2025dynamicface, luo2025canonswap}.

Traditional face swapping techniques primarily rely on geometric warping and blending, which often introduces visible artifacts and lacks realism~\cite{dhanyalakshmi2025survey}. The advent of deep learning, particularly Generative Adversarial Networks (GANs), revolutionized face swapping by enabling higher-quality synthesis~\cite{kang2025zero, moussa2025face}. However, GAN-based methods frequently require auxiliary modules like facial masks and blending stages, which limit generalizability and struggle with diverse poses and lighting conditions~\cite{luo2025canonswap}. More importantly, most of these methods operate frame-by-frame without modeling temporal relationships, leading to instability in video-based face swapping~\cite{wang2025dynamicface, luo2025canonswap}.

Recently, diffusion probabilistic models have emerged as a promising alternative to GANs. Denoising Diffusion Probabilistic Models (DDPMs) \cite{ho2020denoising} and Denoising Diffusion Implicit Models (DDIMs) \cite{song2020denoising} offer improved stability and sample diversity while accelerating generation by 10--50$\times$. Diffusion-based methods like REFace \cite{baliah2025reface} and DiffSwap \cite{diffswap2023} have demonstrated strong results in image-based face swapping by formulating the task as conditional inpainting guided by identity features and facial landmarks. However, directly extending these approaches to video leads to flickering and temporal inconsistencies (see Fig. ~\ref{fig:Video comparisons}) due to the stochastic nature of the diffusion process.
Recent advances in video synthesis with diffusion models have begun to tackle key challenges in temporal coherence and semantic control. Lumiere \cite{lumiere2024} introduces space-time U-Nets to ensure frame-to-frame consistency, while Go-with-the-Flow \cite{burgert2025go} leverages optical flow and warping techniques to maintain temporal alignment. AnyV2V \cite{ku2024anyv2vtuningfreeframeworkvideotovideo} enables zero-shot video generation by editing a single frame and propagating changes temporally without fine-tuning. Despite these developments, existing video face-swapping methods like FaceOff \cite{agarwal2023faceoff} still depend on a full source video, which limits their flexibility and practicality.

In this paper, we introduce a novel diffusion-based framework for high-fidelity, temporally consistent video face swapping. Our method introduces several key contributions: \textbf{(1) Target Structure Guidance:} We use plug-and-play attention injection to better align the structural features from the target frame to the generation. \textbf{(2) Frequency Spectrum Attention Interpolation:} We propose to initialize with DDIM-inverted noise and inject features from the source identity to guide generation and preserve key identity characteristics. \textbf{(3) Flow-guided Attention Temporal Smoothening:} We observe that simply applying an optical flow warping mechanism at the attention level to reduce frame-to-frame flicker and ensure smooth motion transitions.  \textbf{(4) Training-Free Video Swapping:} To our knowledge, ours is the first approach that generalizes image-based face swapping models to the video domain without retraining.
Our experiments demonstrate state-of-the-art performance across multiple metrics, including identity preservation, expression fidelity, temporal coherence, and overall visual quality. Moreover, our method supports one-shot video face swapping with minimal reference data, eliminating the need for extensive fine-tuning. A visual overview of our results is shown in the teaser (Fig.~\ref{fig:teaser}).

\section{Related Work}

\noindent\textbf{Face Swapping Techniques:} Early face swapping approaches used 2D geometric warping and 3D Morphable Models (3DMMs) for facial alignment and blending~\cite{bitouk2008face, blanz2023morphable, zeng2023flowface}. While these methods achieve basic identity transfer, they struggle with realism under varying lighting, expressions, and poses. With the rise of GANs, methods like MegaFS \cite{zhu2022shotfaceswappingmegapixels}, SimSwap \cite{chen2020simswap}, FaceShifter \cite{li2019faceshifter}, and HifiFace \cite{wang2021hififace} emerged. SimSwap introduced identity feature injection via feature matching losses; FaceShifter used a two-stage network for handling occlusions; and HifiFace leveraged 3D priors for better geometry control.
However, these approaches are mostly limited to static images. When naively extended to videos by frame-wise inference, they introduce severe flickering and loss of temporal coherence due to the lack of temporal modeling ability.

\noindent\textbf{Video Based Face Swapping:}  
FaceOff~\cite{agarwal2023faceoff} focuses on source video to target video face swapping, requiring a full source video to edit the target sequence. In contrast, these methods do not address the more challenging source image to target video setting, where only a single source image is available. To this end, we propose a training-free, diffusion-based approach that adapts existing image-based diffusion face swapping methods to videos. To the best of our knowledge, this is the first work tackling this problem.

\noindent\textbf{Diffusion-Based Face Swapping:} Diffusion models have shown impressive results in image generation, offering better sample diversity and stability than GANs. In face swapping, methods such as DiffFace \cite{kim2022diffface} employ ID-conditional DDPMs to inpaint masked regions using source identity features. DiffSwap \cite{diffswap2023} improves control by incorporating 3DMM-aware masks and expression alignment. REFace\cite{baliah2025reface} further improves the performance of face swapping by training the diffusion model as an in-paint task with multistep DDIM sampling at training to improve identity transferability. 
FaceAdapter \cite{han2024faceadapterpretraineddiffusion} adapts pre-trained face generation models to specific tasks by refining facial features and improving realism. It enhances generation quality across tasks without requiring full model retraining. Although effective in these aspects, its performance can still be limited in highly dynamic or complex scenarios, leaving room for further improvement.

Despite these advances, existing diffusion-based methods are designed for static images. When extended to videos, the lack of temporal awareness leads to inconsistencies and identity drift. Moreover, most methods require extensive training or pretraining, and iterative denoising remains computationally expensive.

\noindent\textbf{Identity-Preserving and Temporally Consistent Generation:} Efforts to improve identity preservation and temporal consistency in diffusion-based generation include IP-Adapter \cite{ye2023ipadaptertextcompatibleimage} and InstantID \cite{wang2024instantidzeroshotidentitypreservinggeneration}, which improve identity fidelity using plug-and-play features. 
However, these models often lack high-frequency detail preservation or fail to generalize across long video sequences. Moreover, existing models rely heavily on text inputs to guide the generation process. This limits their applicability to face-swapped video generation, which depends on a reference image rather than textual prompts. As a result, purpose-built architectures that explicitly integrate both identity preservation and temporal consistency remain largely underexplored.

\noindent\textbf{Video Generation with Diffusion Models:} Recent developments in video diffusion modeling show potential for temporally consistent generation. Lumiere \cite{lumiere2024} employs a space-time U-Net to generate realistic videos with coherent motion. AnyV2V \cite{ku2024anyv2vtuningfreeframeworkvideotovideo} and I2VEdit \cite{ouyang2024i2veditfirstframeguidedvideoediting}, achieve zero-shot video editing by modifying only the first frame and propagating changes without retraining. Go-with-the-Flow \cite{burgert2025go} uses optical flow and warping techniques followed by a small video-specific training to maintain temporal alignment accross frames.
Despite showing some progress, most existing pipelines are not designed for identity-preserving face swapping. They suffer from flickering and identity drift, highlighting the need for a specialized architecture for video face swapping.





\section{Preliminaries}

\noindent \textbf{Problem Setting:} Let the target video frames be denoted as \(\{x^{\text{tar}}_i\}_{i=1}^n\), where \(n\) is the total number of frames, and let \(x^{\text{src}}\) represent the source image. The objective of video face swapping is to transfer the identity characteristics from \(x^{\text{src}}\) onto each frame of the target video, while preserving the original pose, expression, and background, resulting in a swapped video sequence \(\{x^{\text{swap}}_i\}_{i=1}^n\). Unlike image-based face swapping methods, the video setting imposes the additional constraint of temporal consistency, requiring the generated frames to be coherent across time and free from flickering artifacts.

\begin{figure*}[htbp]
    \centering
    \includegraphics[width=\textwidth]{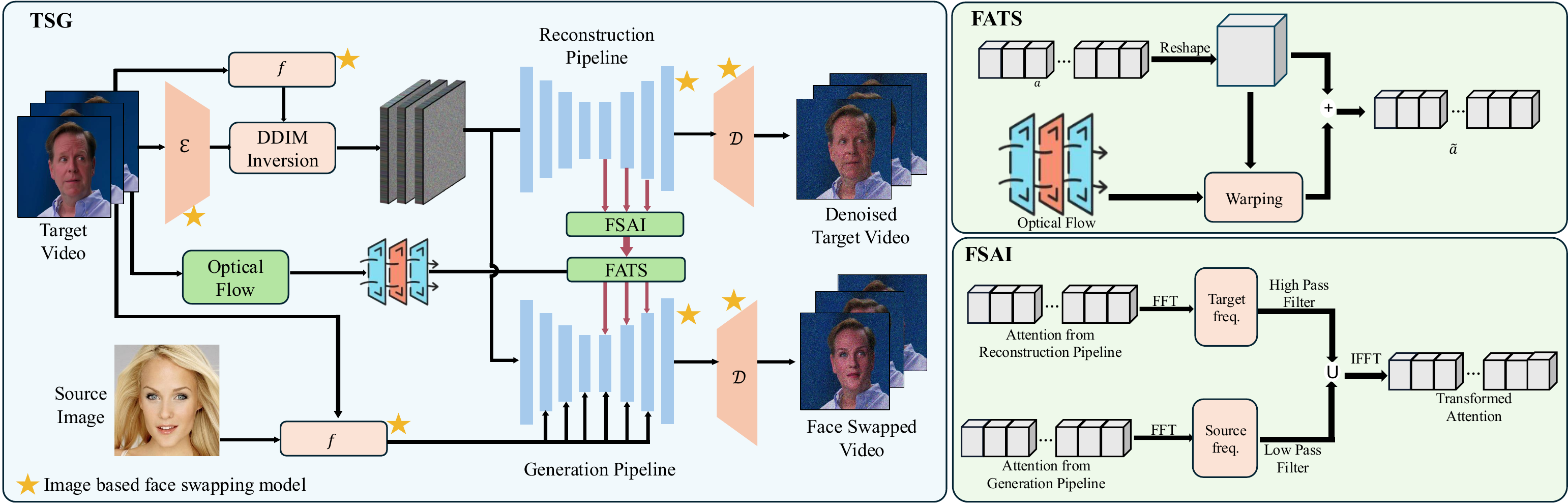}
    \caption{\textbf{VFace overview with its key modules for video face swapping.} The pipeline consists of three core components: (1) Target Structure Guidance (TSG), which aligns structural features from the target video to guide generation; (2) Frequency Spectrum Attention Interpolation (FSAI), which performs attention feature blending in the frequency domain to decouple identity and structure cues; and (3) Flow-guided Attention Temporal Smoothening (FATS), which ensures temporal coherence by propagating attention features across frames using optical flow. Together, these modules enable identity-preserving, structure-aware, and temporally consistent face-swapping.}
    \label{fig:architecture}
\end{figure*}

\noindent \textbf{Diffusion and DDIM inversion:} 
Denoising diffusion models (DDPMs) define a forward Markov chain that gradually adds Gaussian noise to data and train a neural network to reverse this process. Song \textit{et al.}~\cite{song2020denoising} introduced DDIMs as a class of non-Markovian diffusion processes that yield the same training objective as DDPMs, but allow a deterministic generative path. In DDIM sampling, one can set the added noise to zero at each step, so that starting from a latent $z_T$, the model deterministically produces a final image $z_0$ (and conversely each $z_0$ is, in principle, uniquely determined by its $z_T$). DDIM inversion \cite{dhariwal2021diffusion} refers to the inverse of this process: given an image $z_0$, one seeks the corresponding initial noise $z_T$ that would generate it. In other words, inversion maps an image back to its noisy latent representation by running the DDIM update steps in reverse.

Mathematically, a DDIM generative step from time $t$ to $t-1$ can be written as a linear combination of the current state and the predicted noise:
\begin{equation}
    z_{t-1} = m_t\,z_t + n_t\,\epsilon_\theta(z_t,t),
\end{equation}
where $m_t, n_t$ are known scalars determined by the chosen noise schedule. For example, one may take $m_t = \frac{1}{\sqrt{\alpha_t}}$ and 
$n_t = \sqrt{1 - \bar{\alpha}_{t-1}} - \frac{\sqrt{1 - \bar{\alpha}_t}}{\sqrt{\alpha_t}}$, in the usual DDIM parameterization.

Inversion proceeds by solving this equation for $x_t$:
\begin{equation}
    z_t = \frac{z_{t-1} - n_t\,\epsilon_\theta(z_t,t)}{m_t}.
\end{equation}
Because $\epsilon_\theta(z_t,t)$ is unknown when inverting (we only know $z_{t-1}$ at that step), we use the \textit{local linearity} approximation $\epsilon_\theta(z_t,t) \approx \epsilon_\theta(z_{t-1},t)$. Substituting this gives the practical inverse update:
\begin{equation}
    z_t \approx \frac{z_{t-1} - n_t\,\epsilon_\theta(z_{t-1},t)}{m_t}.
\end{equation}

Starting from an observed image \( z_0 \), DDIM inversion performs a deterministic reverse diffusion process by iteratively applying update steps from \( t = 1 \) to \( T \), yielding a latent variable \( z_T \) that theoretically maps back to the original image under the forward diffusion process. While this inversion is designed to reconstruct \( z_0 \), it also facilitates practical applications such as \textit{image editing}~\cite{Mokady_2023_CVPR, tumanyan2023plug}. Specifically, one can initialize the denoising process with the inverted latent \( z_T \) of a source image (e.g., a cat) and condition the pretrained diffusion model on an alternative text prompt (e.g., ``a photo of a dog''). The resulting sample retains the structural and compositional characteristics of the source image while reflecting the semantics of the new conditioning input.

\noindent\textbf{Diffusion-based Image Face Swapping.}
Our method is based on existing diffusion-based image face swapping approach, ~\cite{baliah2025reface, diffswap2023} which typically use a conditioning function $f$ that incorporates target-specific information, such as landmarks and facial masks, along with identity features from the source image. These conditions are injected via cross-attention mechanisms during generation. Additionally, DDIM sampling is initialized with random Gaussian noise, following standard diffusion procedures. Without loss of generality, we demonstrate the effectiveness of our method using the REFace model~\cite{baliah2025reface} as the baseline.

\section{VFace}

We propose a lightweight and effective framework called \textit{VFace} for consistent and realistic face swapping in videos. Our method builds on a pretrained diffusion model and introduces three key components: (i) target structure guidance (sec. \ref{TSG}), (ii) frequency spectrum attention interpolation for identity preservation (sec. \ref{FSAI}), and (iii) flow-guided attention temporal smoothening (sec. \ref{FATS}) to mitigate temporal flickering.

\subsection{Target Structure Guidance (TSG)}
\label{TSG}

Given a sequence of target video frame latents $\{\mathbf{z}^{\text{tar}}_i\}_{i=1}^T$, we first compute the DDIM-inverted noise representations $\{\mathbf{I}^{\text{tar}}_i\}_{i=1}^T$ for the target video using deterministic DDIM inversion \cite{song2020denoising}. These noise vectors approximate the latent noise that would reconstruct the original video frame latents ${\mathbf{z}^{\text{tar}}_i}$ through the forward denoising process, assuming the same conditioning used during inversion.

Following the plug-and-play editing paradigm \cite{tumanyan2023plug}, we adopt a dual-branch architecture consisting of a \textit{reconstruction pipeline} and a \textit{generation pipeline} (Fig.~\ref{fig:architecture}). The reconstruction pipeline denoises $\mathbf{I}^{\text{tar}}_i$ to recover $\mathbf{z}^{\text{tar}}_i$, providing the intermediate attention maps (queries $\mathbf{q}_t^i$ and keys $\mathbf{k}_t^i$)  to the generation pipeline to facilitate target video's structure guidance to generation pipeline at each timestep $t$ for every frame $i$. 
To guide generation, the same target noise $\mathbf{I}^{\text{tar}}$ is denoised in the generation pipeline with the swapping conditions, where we inject attention features from the target’s reconstruction process. Specifically, at each denoising step $t$, we replace the query and key tensors of the generation pipeline with those from the corresponding timestep in the reconstruction pipeline:
\begin{equation}
\mathbf{q}_t^{\text{gen}} \leftarrow \mathbf{q}_t^{i}, \quad \mathbf{k}_t^{\text{gen}} \leftarrow \mathbf{k}_t^{i}.
\end{equation}

This plug-and-play attention injection transfers pose and structure from the target frame while allowing the source identity to be synthesized in the generation pipeline. It ensures spatial consistency across frames and eliminates the need for model retraining or finetuning.

\subsection{Frequency Spectrum Attention Interpolation (FSAI)}
\label{FSAI}

While TSG preserves facial structure and pose, it can diminish the influence of the source identity due to the dominance of DDIM-inverted noise from the target frames and the attention feature replacements. To address this, we introduce the FSAI mechanism that selectively blends the source identity features from the generation pipeline with structural features from the reconstruction pipeline in the frequency domain. 

\begin{figure}[!htb]
    \centering
    \includegraphics[width=1.0\linewidth]{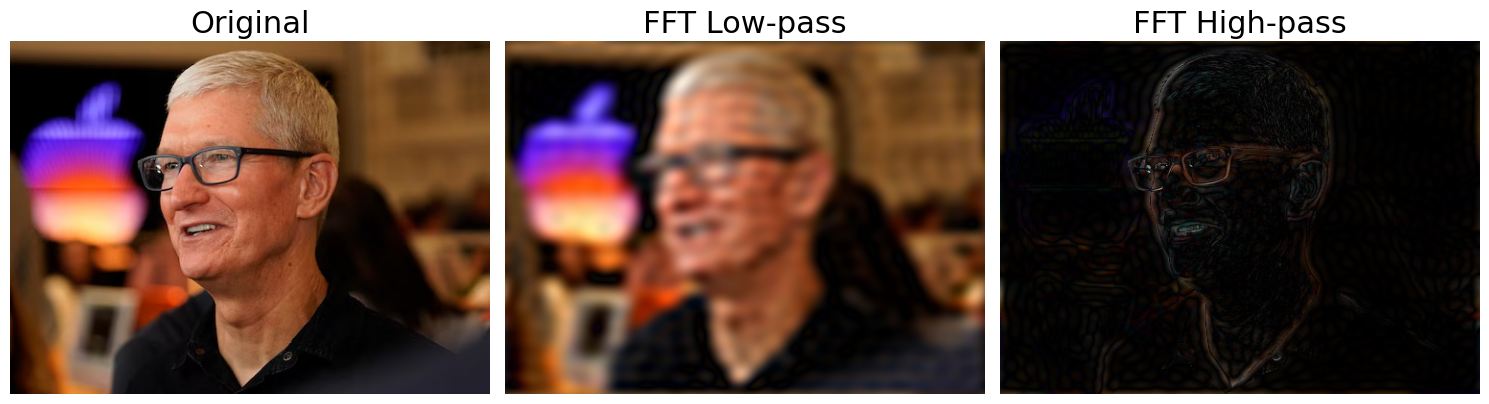}
    \caption{Low pass and High pass filtered source images.}
    \vspace{-0.3em}
    \label{fig:FFT_frequency}
    \vspace{-1em}
\end{figure}

\begin{figure}
    \centering
    \includegraphics[width=1\linewidth]{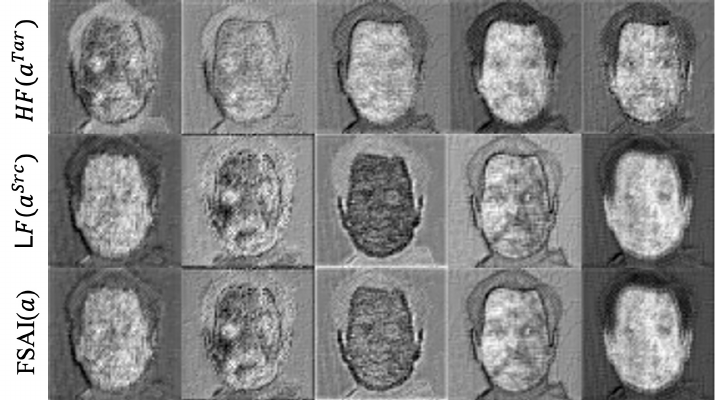}
    \caption{Visualization of the first five channels of the $q$ vector from our FSAI module at the 40th DDIM step, reshaped into square maps.}
    \label{fig:attention_visualization}
\end{figure}

We observe that the low-frequency components of images encode coarse semantic information, such as facial identity and overall appearance, while the high-frequency components capture localized details and structural cues, such as facial hair and fine textures, as illustrated in Figure~\ref{fig:FFT_frequency}. However, directly applying frequency interpolation between the target video frames and the source image does not yield satisfactory results, as it fails to account for the differences in identity and spatial positioning between the source and target. This limitation motivates our approach of performing frequency interpolation on the query (q) and key(k)  vectors of the attention mechanism instead. As illustrated in Figure~\ref{fig:attention_visualization}, these vectors effectively capture the spatial structure of the input, making them a more suitable representation for aligning identity and pose across frames.


Let $\mathbf{a}^{\text{src}}$ and $\mathbf{a}^{\text{tar}}_i$ denote the attention features—either queries ($\mathbf{q}$) or keys ($\mathbf{k}$)—from the source-guided generation pipeline and the reconstruction pipeline of the $i$-th target frame, respectively. Instead of interpolating these features directly in the spatial domain, we perform the interpolation in the frequency domain, which enables a cleaner separation between identity and structural cues. 

We apply a one-dimensional Fast Fourier Transform (FFT) along the temporal or channel-wise dimension of each attention tensor. We then split the resulting spectrum using a ratio $\rho$ (e.g., $\rho = 0.8$), and construct a hybrid spectrum by combining the low-frequency components from the generation pipeline with the high-frequency components from the reconstruction pipeline. An inverse FFT is then applied to obtain interpolated attention maps in the original domain:
\begin{equation}
    \text{FSAI}(\mathbf{a}^{\text{src}},\mathbf{a}^{\text{tar}}_i) = \text{IFFT}\left(
        \text{FFT}(\mathbf{a}^{\text{tar}}_i)_{\text{high}} \cup \text{FFT}(\mathbf{a}^{\text{src}})_{\text{low}}
    \right),
\end{equation}
where $\text{FSAI}(\mathbf{a}^{\text{src}},\mathbf{a}^{\text{tar}}_i) $ is the final interpolated attention tensor used in the generation pipeline.

This frequency-aware attention fusion ensures that the coarse identity cues from the source are preserved, while still benefiting from the structural fidelity and alignment of the target's motion. The mechanism is simple, training-free, and integrates seamlessly into the dual-branch architecture. In practice, we apply this interpolation to both query and key tensors of cross-attention layers at each denoising timestep and across all frames:
\begin{align}
    \mathbf{q}_t^{\text{gen}} &\leftarrow \text{FSAI}(\mathbf{q}_t^{\text{src}}, \mathbf{q}_t^i), \\
    \mathbf{k}_t^{\text{gen}} &\leftarrow \text{FSAI}(\mathbf{k}_t^{\text{src}}, \mathbf{k}_t^i).
\end{align}

By leveraging frequency decomposition, our method provides a principled means of identity–structure disentanglement in the attention space, which results in both high-ID transferability and temporally consistent video face swaps.

\subsection{Flow-guided Attention Temporal Smoothening (FATS)}
\label{FATS}

Despite structural consistency across frames, the inherently stochastic nature of diffusion models leads to subtle flickering and temporal artifacts during video generation~\cite{chen2025temporal}. Prior approaches mitigate this using spatiotemporal attention layers~\cite{wang2025swap, wang2023videofactory, zhou2024upscale}, but these typically require extensive training on large-scale video datasets and often restrict inference to short sequences.

We propose a simple, training-free, inference-time solution called \textit{Flow-guided Attention Temporal Smoothening (FATS)}. Unlike Go-With-the-Flow~\cite{burgert2025go}, which performs optical flow-based warping at the initial noise level, FATS operates at the attention level. Specifically, it warps attention maps using optical flow computed from a downsampled version of the target video (e.g., $64 \times 64$ resolution), promoting temporal coherence during early DDIM steps.

Let $x_{\text{prev}} \in \mathbb{R}^{B \times C \times H \times W}$ denote intermediate features (e.g., query/key) across $B$ frames, and let $\mathbf{f} \in \mathbb{R}^{(B{-}1) \times 2 \times H \times W}$ be the optical flow between consecutive frames. For each frame $i$, we warp $x_{\text{prev}}[i]$ toward frame $i{+}1$ using $\mathbf{f}[i]$, and blend it with the unwarped representation of frame $i{+}1$:

\begin{equation}
x_{i+1}^{\text{aligned}} = \alpha \cdot x_{i+1} + (1 - \alpha) \cdot \text{Warp}(x_i, \mathbf{f}_i)
\end{equation}

\noindent where $\alpha \in [0,1]$ controls the smoothness of transitions. We set $\alpha = 0.8$ in our experiments. This alignment is applied during the first $T_1$ DDIM steps (e.g., $T_1 = 10$), when attention maps are at $64 \times 64$ resolution and follow Frequency Spectrum Attention Interpolation.

FATS has several advantages over prior flow-based strategies: (i) it requires no fine-tuning or retraining of the diffusion model, (ii) it provides controllability by targeting specific attention layers and sampling steps, and (iii) it is less sensitive to flow estimation errors due to its soft blending formulation. FATS is a plug-and-play module that significantly enhances temporal consistency (Table ~\ref{tab:ablation}) by enforcing smooth transitions in the attention space.

\section{Experiments and Results}
\noindent \textbf{Datasets:}
For source images, we use the CelebA \cite{liu2015faceattributes} dataset, containing over 200K celebrity images. For the target videos, we utilize the VFHQ \cite{xie2022vfhq} test set comprising 50 videos, and the CelebV-HQ \cite{zhu2022celebvhqlargescalevideofacial} dataset, which includes approximately 100 videos ranging from 3 to 20 seconds in duration at a resolution of 512 × 512. Following the ReFace protocol for image-based face swapping comparisons, we employ both CelebA and FFHQ \cite{karras2019style}, the latter containing around 70,000 high-quality images at 512 × 512 resolution.

\begin{figure*}
  \centering
  \animategraphics[width=0.9\linewidth,loop,poster=first]{10}{Videos/compare_video/frame_0000}{01}{24}
  \caption{Video comparison with REFace as baseline on VFHQ dataset. Play the video with adobe reader}
  \label{fig:Video comparisons}
\end{figure*}

\begin{figure*}
    \centering
    \includegraphics[width=\linewidth]{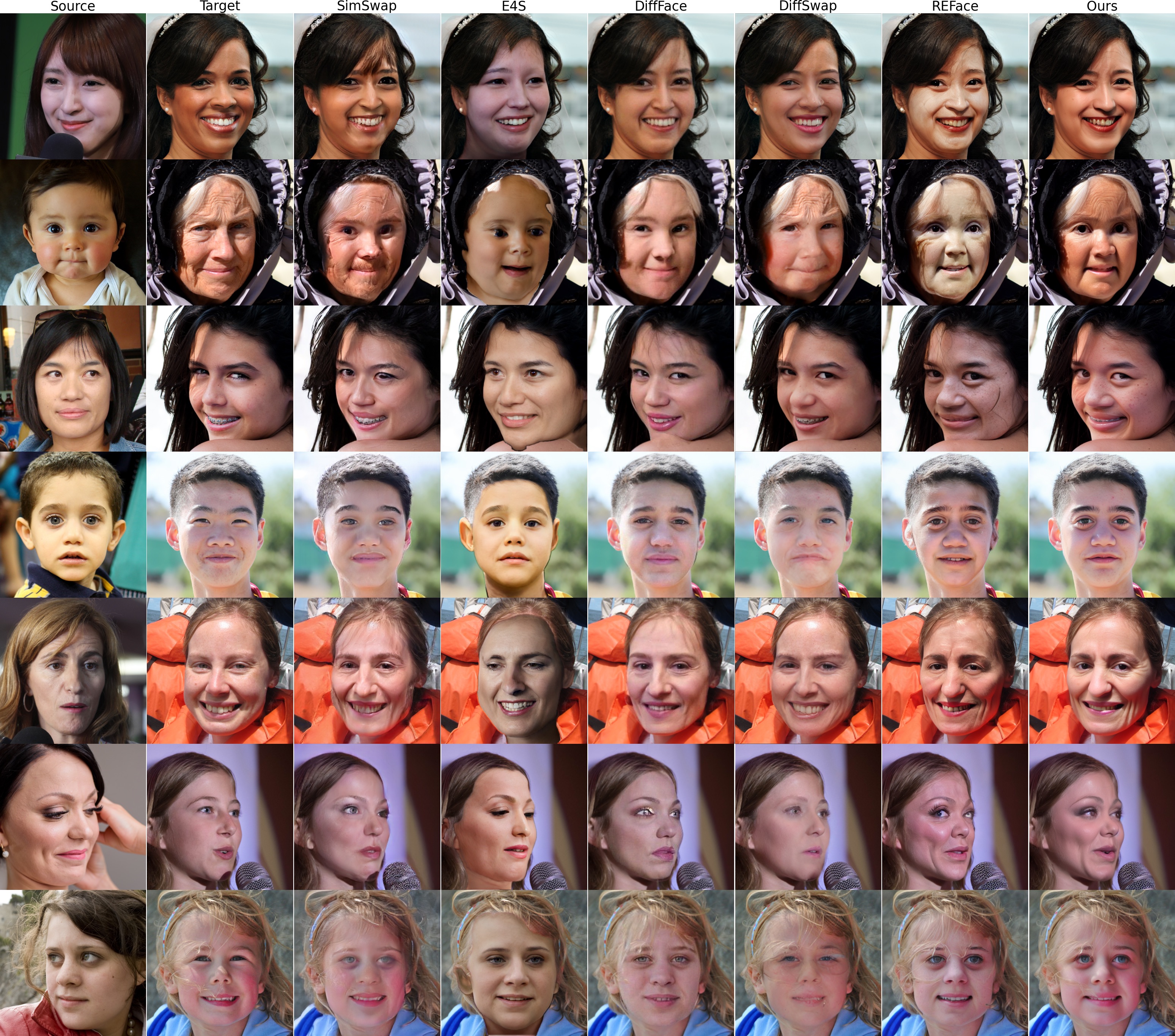}
    \caption{Qualitative comparison on FFHQ dataset.}
    \vspace{-0.2em}
    \label{fig:FFHQ compare}
    \vspace{-1em}
\end{figure*}

\noindent \textbf{Evaluation metrics:}
To evaluate video-based face swapping, we follow the protocol of REFace \cite{baliah2025reface}. We compute the Fréchet Inception Distance (FID) \cite{heusel2017gans} between the generated frames and images from the CelebV-HQ \cite{zhu2022celebvhqlargescalevideofacial} dataset to assess image quality. For pose and expression accuracy, we employ HopeNet \cite{doosti2020hopenetgraphbasedmodelhandobject} and Deep3DFaceRecon \cite{deng2020accurate3dfacereconstruction}, respectively. To measure identity preservation, we use ArcFace\cite{Deng_2022} to extract facial identity features and compare them against a gallery of 1,000 source images from , reporting Top-1 and Top-5 retrieval accuracy. To evaluate temporal consistency and overall video fidelity, we adopt FVD \cite{unterthiner2019accurategenerativemodelsvideo} and Content-Debiased FVD (CD-FVD) \cite{ge2024contentbiasfrechetvideo}.

\noindent \textbf{Implementation details:}
We implement our method over a recent diffusion-based face swapping framework REFace~\cite{baliah2025reface}. To extract the structural information from the target video, we perform DDIM inversion using 50 steps, which provides an efficient and effective latent representation of the target frames. This is notably faster than the 500-step inversion used in methods like AnyV2V, while still maintaining high-quality results. We process the video in batches of 6 frames, using a sliding window approach where the last frame of the previous batch is reused to maintain temporal consistency across batches. We apply our attention manipulations in all cross-attention layers on the output side (upsampling) of the denoising U-Net, and use RAFT-Large \cite{zhang2024raftadaptinglanguagemodel} to compute the optical flow at $64 \times 64$ resolution.  All experiments are conducted on a 40GB A100 GPU.

\noindent \textbf{Image Based Results:}
In video-based face swapping, there is a critical trade-off between accurately retrieving the identity of the target person and effectively transferring the pose and expression from the source video. Maintaining temporal consistency across frames is especially important, as frame-to-frame variations in these factors can lead to visual artifacts and instability. Since most existing face swapping methods are image-based, we first evaluate our video-based approach in the image domain w.r.t FID, ID retrieval, Pose and Expression to enable a fair and consistent comparison. 

Table~\ref{tab:FFHQ} demonstrates that VFace outperforms all other image-based models across all evaluation metrics in FFHQ dataset, with ReFace being a close second. This is further supported by the qualitative results on the FFHQ dataset, shown in Figure~\ref{fig:FFHQ compare}. Our model more accurately preserves the identity and fine-grained details of the target image while closely following the pose and expression of the target as well. In Table~\ref{tab:CelebA} VFace shows better FID on CelebA dataset, while having comparable results in terms of ID and Pose to previous SOTA. 

\begin{table}[!htbp]
\centering
\small
\setlength{\tabcolsep}{3pt}
\begin{tabular}{lccccc}
\toprule
\multirow{2}{*}{\textbf{Method}}  & \multirow{2}{*}{\textbf{FID}$\downarrow$}    &   \multicolumn{2}{c}{\textbf{ID retrieval $\uparrow$}}   & \multirow{2}{*}{\textbf{Pose}$\downarrow$}   &  \multirow{2}{*}{\textbf{Expr.}$\downarrow$}    \\ 
 &&   Top-1 &  Top-5 \\
\midrule
MegaFS \cite{zhu2021one} &12.00 & 59.60\% & 74.10\% & 3.33 & 1.11  \\
HifiFace  \cite{wang2021hififace} &11.58 & 75.30\% & 87.1\% & 3.28 & 1.41 \\
SimSwap\cite{chen2020simswap} &13.80 & 90.60\% & 96.40\% & 2.98 & 1.07  \\
E4S\cite{liu2023fine} &12.38 & 70.20\% & 82.73\% & 4.50 & 1.31  \\
DiffFace \cite{kim2022diffface} & 8.59 & 87.20\% & 94.40\% & 3.80 & 2.28  \\
DiffSwap\cite{zhao2023diffswap} & 8.58 & 78.20 \% & 93.60 \% & \underline{2.92} & 1.10 \\
REFace \cite{baliah2025reface}&\underline{5.53}& \underline{95.40\%} & \textbf{98.70\%} & 3.74 & \underline{1.04}  \\
VFace (Ours) &\textbf{4.29}& \textbf{96.20\%} & \underline{98.40\% }& \textbf{2.90} & \textbf{0.86}  \\
\bottomrule
\end{tabular}
\vspace{-0.5em}
\caption{Comparison on FFHQ dataset.} 

\label{tab:FFHQ}
\end{table}

\begin{table}[!htbp]
\centering
\small
\setlength{\tabcolsep}{3pt}
\begin{tabular}{lccccc}
\toprule
\multirow{2}{*}{\textbf{Method}}  & \multirow{2}{*}{\textbf{FID}$\downarrow$}    &   \multicolumn{2}{c}{\textbf{ID retrieval $\uparrow$}}   & \multirow{2}{*}{\textbf{Pose}$\downarrow$}   &  \multirow{2}{*}{\textbf{Expr.}$\downarrow$}    \\ 
 &&   Top-1 &  Top-5 \\
\midrule
MegaFS \cite{zhu2021one} &20.3 & 73.9\% & 80.3\% & 5.83 & 1.20  \\
HifiFace  \cite{wang2021hififace} &12.68 & 88.30\% & 94.30\% & 2.92 & 1.09 \\
SimSwap\cite{chen2020simswap} &10.7 & 95.30\% & \underline{98.60\%} & 2.89 & \underline{0.99}  \\
E4S\cite{liu2023fine} &14.58 & 83.30\% & 90.80\% & 4.15 & 1.16  \\
DiffFace \cite{kim2022diffface} &10.29 & 94.6\% & 97.9\% & 36.1 & 2.15  \\
DiffSwap\cite{zhao2023diffswap} & 9.16 & 75.6 \% & 90.9 \% & \textbf{2.48} & \textbf{0.77} \\
REFace \cite{baliah2025reface}&\underline{6.09 }& \textbf{98.8\%} & \textbf{99.6\%} & 3.51 & 0.96  \\
    VFace (Ours) &\textbf{5.42}& \underline{96.2\%} & 98.4\% & \underline{2.63 }& \textbf{0.77}  \\
\bottomrule
\end{tabular}
\vspace{-0.5em}
\caption{Comparison on CelebA dataset.} 

\label{tab:CelebA} \vspace{-1em}
\end{table}



\begin{table}[!htp]
\centering

\setlength{\tabcolsep}{1pt}
\scalebox{0.7}{
\begin{tabular}{lccccccc}
\toprule
\multirow{2}{*}{\textbf{Method}}  & \multirow{2}{*}{\textbf{CD-FVD}$\downarrow$} & \multirow{2}{*}{\textbf{FVD}$\downarrow$}    &   \multirow{2}{*}{\textbf{ID Sim.}$\uparrow$}  & \multicolumn{2}{c}{\textbf{ID retrieval $\uparrow$}}   & \multirow{2}{*}{\textbf{Pose}$\downarrow$}   &  \multirow{2}{*}{\textbf{Expr.}$\downarrow$}    \\ 
 &&&&   Top-1 &  Top-5 \\
\midrule
Vanilla REFace\cite{baliah2025reface}& 211.27 & \underline{204.53} & \underline{0.55}  & \underline{96.30} & \textbf{97.18} & \underline{4.58} & \underline{1.15}\\
\cite{baliah2025reface} + AnyV2V& 233.04 & 281.05 & 0.36  & 72.18  & 78.35& 6.87 &1.41\\
\cite{baliah2025reface} + Go-with-the-flow& \underline{190.00} & 302.10 & 0.43  & 77.41 & \underline{80.52} & 7.21 &1.98\\
\cite{baliah2025reface} + VFace(Ours)& \textbf{163.82} & \textbf{159.99 }& \textbf{0.58}   & \textbf{96.54}  & \textbf{97.18} & \textbf{4.13 }& \textbf{1.08} \\
\bottomrule
\end{tabular}}
\vspace{-0.5em}
\caption{Comparison on CelebVHQ dataset. }
\label{tab:CelebVHQ}\vspace{-1em}
\end{table}


\noindent \textbf{Video Based Results:}
By closely tracking the pose and expression of the target video on a frame-by-frame basis and integrating optical flow into the attention features, our method achieves strong temporal consistency. As reported in Table~\ref{tab:CelebVHQ} and Table~\ref{tab:VFHQ}, our model consistently outperforms all baselines in terms of CD-FVD and FVD, including combinations of REFace with various video generation approaches, on both the VFHQ and CelebVHQ datasets. Qualitative comparisons in Figure~\ref{fig:Video comparisons} further demonstrate the superior temporal coherence and effective identity preservation achieved by our method. 

\begin{table}[!htp]
\centering

\setlength{\tabcolsep}{1pt}
\scalebox{0.7}{
\begin{tabular}{lccccccc}
\toprule
\multirow{2}{*}{\textbf{Method}}  & \multirow{2}{*}{\textbf{CD-FVD}$\downarrow$} & \multirow{2}{*}{\textbf{FVD}$\downarrow$}    &   \multirow{2}{*}{\textbf{ID Sim.}$\uparrow$}  & \multicolumn{2}{c}{\textbf{ID retrieval $\uparrow$}}   & \multirow{2}{*}{\textbf{Pose}$\downarrow$}   &  \multirow{2}{*}{\textbf{Expr.}$\downarrow$}    \\ 
 &&&&  Top-1 &  Top-5 \\
\midrule
Vanilla REFace\cite{baliah2025reface}& 218.01& \underline{232.09} & \textbf{0.58}   & \textbf{99.90} & \textbf{100.0} & 3.50 & \underline{1.20}\\
\cite{baliah2025reface} + AnyV2V& 273.62 & 261.61 & 0.46 & 96.15 &98.54 &\underline{3.37}  &1.31\\
\cite{baliah2025reface} + Go-With-The-FLow& \underline{197.10}& 315.14  & 0.54  & 97.92& 99.38 & 5.05 & 1.56 \\
\cite{baliah2025reface} + VFace(Ours)& \textbf{170.90} & \textbf{230.31} & \underline{0.57} & \underline{99.58}  & \textbf{100.0} & \textbf{2.70} & \textbf{1.06} \\
\bottomrule
\end{tabular}}
\caption{Comparison on VFHQ  video test set.} 
\label{tab:VFHQ}
\end{table}

\noindent\textbf{Efficiency:} We compare the per frame inference time with other video models in Table~\ref{tab:time_comparison}. Our method adds 5.3s overhead compared to vanilla REFace due to the DDIM inversions. Notably, ours provides a much faster inference compared to AnyV2V and Go-with-the-flow.

\begin{table}[!htbp]
\centering
\setlength{\tabcolsep}{1pt}
\small
\begin{tabular}{l|c| c| c|c}
\toprule
\textbf{Method} & REFace & Go-with-the-flow & AnyV2V & Ours (VFace) \\
\midrule
\textbf{Time (s)} & 4.8 & 16.7 & 15.8 & 10.1 \\
\bottomrule
\end{tabular}
\caption{Inference time comparison (seconds per frame).}
\label{tab:time_comparison}
\end{table}

\noindent \textbf{Ablation:} We evaluate the contribution of each of our three core components to video face swapping on the VFHQ dataset, as shown in Table~\ref{tab:ablation}. Incorporating TSG significantly improves pose fidelity. Adding FSAI further enhances identity similarity while retaining the pose improvements introduced by TSG. Finally, FATS boosts temporal consistency, as reflected in the improved CD-FVD scores.

\begin{table}[!htp]
\centering
\small
\setlength{\tabcolsep}{3pt}
\begin{tabular}{ccc|ccc}
\toprule
\textbf{TSG} & \textbf{FSAI} & \textbf{FATS} & \textbf{CD-FVD} $\downarrow$ & \textbf{ID Sim} $\uparrow$ & \textbf{Pose} $\downarrow$ \\
\midrule
          --   &  --            &        --      & \underline{218.01}  & \textbf{0.58}   & 3.50 \\
\checkmark   &        --       &      --         &  229.09 & 0.53  & \textbf{2.61}  \\
\checkmark   & \checkmark   &       --       &  232.20 & \underline{0.57}  & 2.77 \\
\checkmark   & \checkmark   & \checkmark   &  \textbf{170.90} &  \underline{0.57} &  \underline{2.70} \\
\bottomrule
\end{tabular}
\caption{Ablation study of key components: Target Structure Guidance (TSG), Frequency Spectrum Attention Interpolation (FSAI), and Flow-guided Attention Temporal Smoothening (FATS). We report CD-FVD ($\downarrow$), ID similarity ($\uparrow$), and Pose error ($\downarrow$) on VFHQ test set.}
\label{tab:ablation}
\end{table}


For additional details— including hyperparameter analysis, further ablations, ethical considerations, limitations, and adaptability to FaceAdapter\cite{han2024faceadapterpretraineddiffusion} — please refer to the appendix .

\section{Conclusion}

We propose a training-free video face swapping framework that can be seamlessly integrated with existing diffusion-based image face swapping models. Our key contributions, including Target Structure Guidance, Frequency Spectrum Noise Interpolation and Flow-Guided Temporal Smoothing, significantly enhance pose alignment, expression consistency, and overall video fidelity compared to vanilla image-based models and first-frame editing video diffusion baselines, while preserving or improving identity transferability. Future work will focus on further improving video realism and addressing challenges such as complex occlusions.


\section*{Acknowledgments}

This material is partly based upon work supported by the Center for Identification Technology Research and the National Science Foundation under Grant Numbers 1841517 and 2413309 at Michigan State University; and in part by the BK21 FOUR Project (Bigdata Research and Education Group for Enhancing Social Connectedness Thorough Advanced Data Technology and Interaction Science Research, 5199990913845), funded by the Ministry of Education (MOE, Korea) and the National Research Foundation of Korea (NRF); and in part by the “Regional Innovation System \& Education (RISE)” through the Seoul RISE Center, funded by the Ministry of Education (MOE) and the Seoul Metropolitan Government (2025-RISE-01-018-01).

{\small
\bibliographystyle{ieee_fullname}
\bibliography{egbib}

\begin{thebibliography}{10}\itemsep=-1pt

\bibitem{agarwal2023faceoff}
Aditya Agarwal, Bipasha Sen, Rudrabha Mukhopadhyay, Vinay~P Namboodiri, and CV Jawahar.
\newblock Faceoff: A video-to-video face swapping system.
\newblock In {\em Proceedings of the IEEE/CVF Winter Conference on Applications of Computer Vision}, pages 3495--3504, 2023.

\bibitem{baliah2025reface}
Sanoojan Baliah, Qinliang Lin, Shengcai Liao, Xiaodan Liang, and Muhammad~Haris Khan.
\newblock Realistic and efficient face swapping: A unified approach with diffusion models.
\newblock In {\em Proceedings of the IEEE/CVF Winter Conference on Applications of Computer Vision (WACV)}, 2025.
\newblock Oral.

\bibitem{bitouk2008face}
Dmitri Bitouk, Neeraj Kumar, Samreen Dhillon, Peter Belhumeur, and Shree~K Nayar.
\newblock Face swapping: automatically replacing faces in photographs.
\newblock In {\em ACM SIGGRAPH 2008 papers}, pages 1--8. 2008.

\bibitem{blanz2023morphable}
Volker Blanz and Thomas Vetter.
\newblock A morphable model for the synthesis of 3d faces.
\newblock In {\em Seminal Graphics Papers: Pushing the Boundaries, Volume 2}, pages 157--164. 2023.

\bibitem{burgert2025go}
Ryan Burgert, Yuancheng Xu, Wenqi Xian, Oliver Pilarski, Pascal Clausen, Mingming He, Li Ma, Yitong Deng, Lingxiao Li, Mohsen Mousavi, et~al.
\newblock Go-with-the-flow: Motion-controllable video diffusion models using real-time warped noise.
\newblock {\em arXiv preprint arXiv:2501.08331}, 2025.

\bibitem{chen2025temporal}
Harold~Haodong Chen, Haojian Huang, Xianfeng Wu, Yexin Liu, Yajing Bai, Wen-Jie Shu, Harry Yang, and Ser-Nam Lim.
\newblock Temporal regularization makes your video generator stronger.
\newblock {\em arXiv preprint arXiv:2503.15417}, 2025.

\bibitem{chen2020simswap}
Renwang Chen, Xuanhong Chen, Bingbing Ni, and Yanhao Ge.
\newblock Simswap: An efficient framework for high fidelity face swapping.
\newblock In {\em Proceedings of the 28th ACM International Conference on Multimedia}, pages 2003--2011, 2020.

\bibitem{Deng_2022}
Jiankang Deng, Jia Guo, Jing Yang, Niannan Xue, Irene Kotsia, and Stefanos Zafeiriou.
\newblock Arcface: Additive angular margin loss for deep face recognition.
\newblock {\em IEEE Transactions on Pattern Analysis and Machine Intelligence}, 44(10):5962–5979, Oct. 2022.

\bibitem{deng2020accurate3dfacereconstruction}
Yu Deng, Jiaolong Yang, Sicheng Xu, Dong Chen, Yunde Jia, and Xin Tong.
\newblock Accurate 3d face reconstruction with weakly-supervised learning: From single image to image set, 2020.

\bibitem{dhanyalakshmi2025survey}
Ramamurthy Dhanyalakshmi, Gabriel Stoian, Daniela Danciulescu, and Duraisamy~Jude Hemanth.
\newblock A survey on face-swapping methods for identity manipulation in deepfake applications.
\newblock {\em IET Image Processing}, 19(1):e70132, 2025.

\bibitem{dhariwal2021diffusion}
Prafulla Dhariwal and Alexander Nichol.
\newblock Diffusion models beat gans on image synthesis.
\newblock {\em Advances in neural information processing systems}, 34:8780--8794, 2021.

\bibitem{doosti2020hopenetgraphbasedmodelhandobject}
Bardia Doosti, Shujon Naha, Majid Mirbagheri, and David Crandall.
\newblock Hope-net: A graph-based model for hand-object pose estimation, 2020.

\bibitem{ge2024contentbiasfrechetvideo}
Songwei Ge, Aniruddha Mahapatra, Gaurav Parmar, Jun-Yan Zhu, and Jia-Bin Huang.
\newblock On the content bias in fr\'echet video distance, 2024.

\bibitem{han2024faceadapterpretraineddiffusion}
Yue Han, Junwei Zhu, Keke He, Xu Chen, Yanhao Ge, Wei Li, Xiangtai Li, Jiangning Zhang, Chengjie Wang, and Yong Liu.
\newblock Face adapter for pre-trained diffusion models with fine-grained id and attribute control, 2024.

\bibitem{heusel2017gans}
Martin Heusel, Hubert Ramsauer, Thomas Unterthiner, Bernhard Nessler, and Sepp Hochreiter.
\newblock Gans trained by a two time-scale update rule converge to a local nash equilibrium.
\newblock {\em Advances in neural information processing systems}, 30, 2017.

\bibitem{ho2020denoising}
Jonathan Ho, Ajay Jain, and Pieter Abbeel.
\newblock Denoising diffusion probabilistic models.
\newblock {\em Advances in Neural Information Processing Systems}, 33:6840--6851, 2020.

\bibitem{kang2025zero}
Taewoong Kang, Sohyun Jeong, Hyojin Jang, and Jaegul Choo.
\newblock Zero-shot head swapping in real-world scenarios.
\newblock In {\em Proceedings of the Computer Vision and Pattern Recognition Conference}, pages 10805--10814, 2025.

\bibitem{karras2019style}
Tero Karras, Samuli Laine, and Timo Aila.
\newblock A style-based generator architecture for generative adversarial networks.
\newblock In {\em Proceedings of the IEEE/CVF conference on computer vision and pattern recognition}, pages 4401--4410, 2019.

\bibitem{kim2022diffface}
Kihong Kim, Yunho Kim, Seokju Cho, Junyoung Seo, Jisu Nam, Kychul Lee, Seungryong Kim, and KwangHee Lee.
\newblock Diffface: Diffusion-based face swapping with facial guidance.
\newblock {\em Pattern Recognition}, 163:111451, 2025.

\bibitem{ku2024anyv2vtuningfreeframeworkvideotovideo}
Max Ku, Cong Wei, Weiming Ren, Harry Yang, and Wenhu Chen.
\newblock Anyv2v: A tuning-free framework for any video-to-video editing tasks, 2024.

\bibitem{li2019faceshifter}
Lingzhi Li, Jianmin Bao, Hao Yang, Dong Chen, and Fang Wen.
\newblock Faceshifter: Towards high fidelity and occlusion aware face swapping.
\newblock {\em arXiv preprint arXiv:1912.13457}, 2019.

\bibitem{liu2023fine}
Zhian Liu, Maomao Li, Yong Zhang, Cairong Wang, Qi Zhang, Jue Wang, and Yongwei Nie.
\newblock Fine-grained face swapping via regional gan inversion.
\newblock In {\em Proceedings of the IEEE/CVF Conference on Computer Vision and Pattern Recognition}, pages 8578--8587, 2023.

\bibitem{liu2015faceattributes}
Ziwei Liu, Ping Luo, Xiaogang Wang, and Xiaoou Tang.
\newblock Deep learning face attributes in the wild.
\newblock In {\em Proceedings of International Conference on Computer Vision (ICCV)}, December 2015.

\bibitem{luo2025canonswap}
Xiangyang Luo, Ye Zhu, Yunfei Liu, Lijian Lin, Cong Wan, Zijian Cai, Shao-Lun Huang, and Yu Li.
\newblock Canonswap: High-fidelity and consistent video face swapping via canonical space modulation.
\newblock {\em arXiv preprint arXiv:2507.02691}, 2025.

\bibitem{Mokady_2023_CVPR}
Ron Mokady, Amir Hertz, Kfir Aberman, Yael Pritch, and Daniel Cohen-Or.
\newblock Null-text inversion for editing real images using guided diffusion models.
\newblock In {\em Proceedings of the IEEE/CVF Conference on Computer Vision and Pattern Recognition (CVPR)}, pages 6038--6047, June 2023.

\bibitem{moussa2025face}
Elmokhtar~Mohamed Moussa, Ioannis Sarridis, Emmanouil Krasanakis, Nathan Ramoly, Symeon Papadopoulos, Ahmad-Montaser Awal, and Lara Younes.
\newblock Face-swapping based data augmentation for id document and selfie face verification.
\newblock In {\em Proceedings of the Winter Conference on Applications of Computer Vision}, pages 1421--1428, 2025.

\bibitem{ouyang2024i2veditfirstframeguidedvideoediting}
Wenqi Ouyang, Yi Dong, Lei Yang, Jianlou Si, and Xingang Pan.
\newblock I2vedit: First-frame-guided video editing via image-to-video diffusion models, 2024.

\bibitem{lumiere2024}
Uriel Singer, Adam Polyak, Shai Sheynin, Shaked Dror, Daniel Fried, Tomer Shalev, Yossi Kalman, Avishai Noy, Aviv Shoshan, Sagie Benaim, et~al.
\newblock Lumiere: A space-time diffusion model for video generation.
\newblock {\em arXiv preprint arXiv:2401.12945}, 2024.

\bibitem{song2020denoising}
Jiaming Song, Chenlin Meng, and Stefano Ermon.
\newblock Denoising diffusion implicit models.
\newblock In {\em International Conference on Learning Representations (ICLR)}, 2021.

\bibitem{tumanyan2023plug}
Narek Tumanyan, Michal Geyer, Shai Bagon, and Tali Dekel.
\newblock Plug-and-play diffusion features for text-driven image-to-image translation.
\newblock In {\em Proceedings of the IEEE/CVF Conference on Computer Vision and Pattern Recognition}, pages 1921--1930, 2023.

\bibitem{unterthiner2019accurategenerativemodelsvideo}
Thomas Unterthiner, Sjoerd van Steenkiste, Karol Kurach, Raphael Marinier, Marcin Michalski, and Sylvain Gelly.
\newblock Towards accurate generative models of video: A new metric \& challenges, 2019.

\bibitem{wang2024instantidzeroshotidentitypreservinggeneration}
Qixun Wang, Xu Bai, Haofan Wang, Zekui Qin, Anthony Chen, Huaxia Li, Xu Tang, and Yao Hu.
\newblock Instantid: Zero-shot identity-preserving generation in seconds, 2024.

\bibitem{wang2025dynamicface}
Runqi Wang, Sijie Xu, Tianyao He, Yang Chen, Wei Zhu, Dejia Song, Nemo Chen, Xu Tang, and Yao Hu.
\newblock Dynamicface: High-quality and consistent video face swapping using composable 3d facial priors.
\newblock {\em arXiv preprint arXiv:2501.08553}, 2025.

\bibitem{wang2023videofactory}
Wenjing Wang, Huan Yang, Zixi Tuo, Huiguo He, Junchen Zhu, Jianlong Fu, and Jiaying Liu.
\newblock Videofactory: Swap attention in spatiotemporal diffusions for text-to-video generation.
\newblock 2023.

\bibitem{wang2025swap}
Wenjing Wang, Huan Yang, Zixi Tuo, Huiguo He, Junchen Zhu, Jianlong Fu, and Jiaying Liu.
\newblock Swap attention in spatiotemporal diffusions for text-to-video generation.
\newblock {\em International Journal of Computer Vision}, pages 1--19, 2025.

\bibitem{wang2021hififace}
Yuhan Wang, Xu Chen, Junwei Zhu, Wenqing Chu, Ying Tai, Chengjie Wang, Jilin Li, Yongjian Wu, Feiyue Huang, and Rongrong Ji.
\newblock Hififace: 3d shape and semantic prior guided high fidelity face swapping.
\newblock In Zhi-Hua Zhou, editor, {\em Proceedings of the Thirtieth International Joint Conference on Artificial Intelligence, {IJCAI-21}}, pages 1136--1142. International Joint Conferences on Artificial Intelligence Organization, 8 2021.
\newblock Main Track.

\bibitem{xie2022vfhq}
Liangbin Xie, Xintao Wang, Honglun Zhang, Chao Dong, and Ying Shan.
\newblock Vfhq: A high-quality dataset and benchmark for video face super-resolution.
\newblock In {\em Proceedings of the IEEE/CVF Conference on Computer Vision and Pattern Recognition}, pages 657--666, 2022.

\bibitem{ye2023ipadaptertextcompatibleimage}
Hu Ye, Jun Zhang, Sibo Liu, Xiao Han, and Wei Yang.
\newblock Ip-adapter: Text compatible image prompt adapter for text-to-image diffusion models, 2023.

\bibitem{zeng2023flowface}
Hao Zeng, Wei Zhang, Changjie Fan, Tangjie Lv, Suzhen Wang, Zhimeng Zhang, Bowen Ma, Lincheng Li, Yu Ding, and Xin Yu.
\newblock Flowface: Semantic flow-guided shape-aware face swapping.
\newblock In {\em Proceedings of the AAAI conference on artificial intelligence}, volume~37, pages 3367--3375, 2023.

\bibitem{zhang2024raftadaptinglanguagemodel}
Tianjun Zhang, Shishir~G. Patil, Naman Jain, Sheng Shen, Matei Zaharia, Ion Stoica, and Joseph~E. Gonzalez.
\newblock Raft: Adapting language model to domain specific rag, 2024.

\bibitem{zhao2023diffswap}
Wenliang Zhao, Yongming Rao, Weikang Shi, Zuyan Liu, Jie Zhou, and Jiwen Lu.
\newblock Diffswap: High-fidelity and controllable face swapping via 3d-aware masked diffusion.
\newblock In {\em Proceedings of the IEEE/CVF Conference on Computer Vision and Pattern Recognition (CVPR)}, pages 11672--11682, 2023.

\bibitem{zhou2024upscale}
Shangchen Zhou, Peiqing Yang, Jianyi Wang, Yihang Luo, and Chen~Change Loy.
\newblock Upscale-a-video: Temporal-consistent diffusion model for real-world video super-resolution.
\newblock In {\em Proceedings of the IEEE/CVF Conference on Computer Vision and Pattern Recognition}, pages 2535--2545, 2024.

\bibitem{diffswap2023}
Yifan Zhou, Yuxuan Li, Wei Wang, Zhiqiang Li, and Dahua Lin.
\newblock Diffswap: Face swapping with diffusion models.
\newblock In {\em Proceedings of the IEEE/CVF Conference on Computer Vision and Pattern Recognition (CVPR)}, 2023.

\bibitem{zhu2022celebvhqlargescalevideofacial}
Hao Zhu, Wayne Wu, Wentao Zhu, Liming Jiang, Siwei Tang, Li Zhang, Ziwei Liu, and Chen~Change Loy.
\newblock Celebv-hq: A large-scale video facial attributes dataset, 2022.

\bibitem{zhu2022shotfaceswappingmegapixels}
Yuhao Zhu, Qi Li, Jian Wang, Chengzhong Xu, and Zhenan Sun.
\newblock One shot face swapping on megapixels, 2022.

\bibitem{zhu2021one}
Yuhao Zhu, Qi Li, Jian Wang, Cheng-Zhong Xu, and Zhenan Sun.
\newblock One shot face swapping on megapixels.
\newblock In {\em Proceedings of the IEEE/CVF conference on computer vision and pattern recognition}, pages 4834--4844, 2021.

\end{thebibliography}
}
\clearpage
\appendix
\section{Appendix}

\subsection{Hyper parameter analysis}

\begin{figure}[!h]
    \centering
    \begin{subfigure}[t]{1\linewidth}
        \centering
        \includegraphics[width=\linewidth]{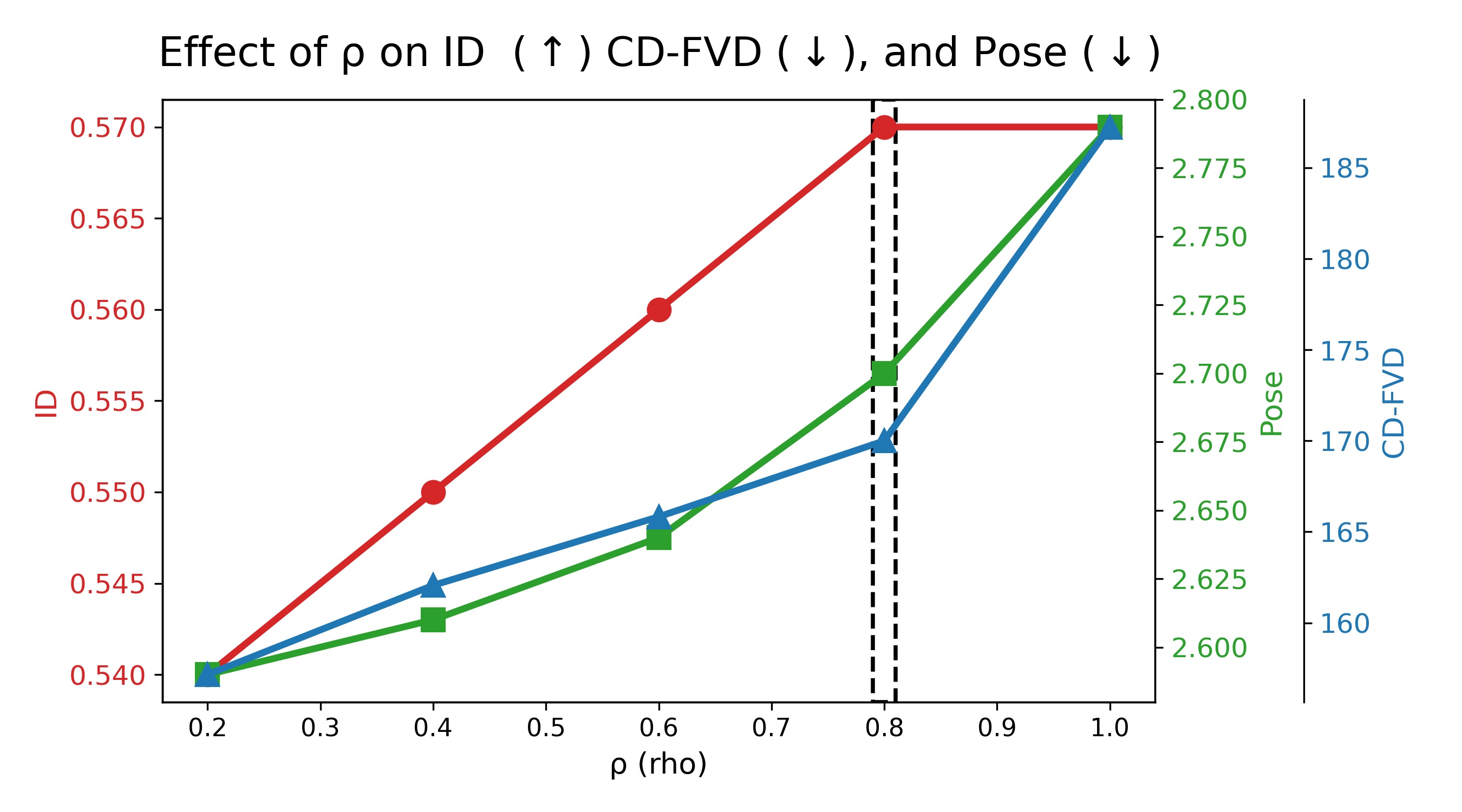}
        \caption{Effect of $\rho$ (frequency split ratio).}
        \label{fig:rho}
    \end{subfigure}
    \hfill
    \begin{subfigure}[t]{1\linewidth}
        \centering
        \includegraphics[width=\linewidth]{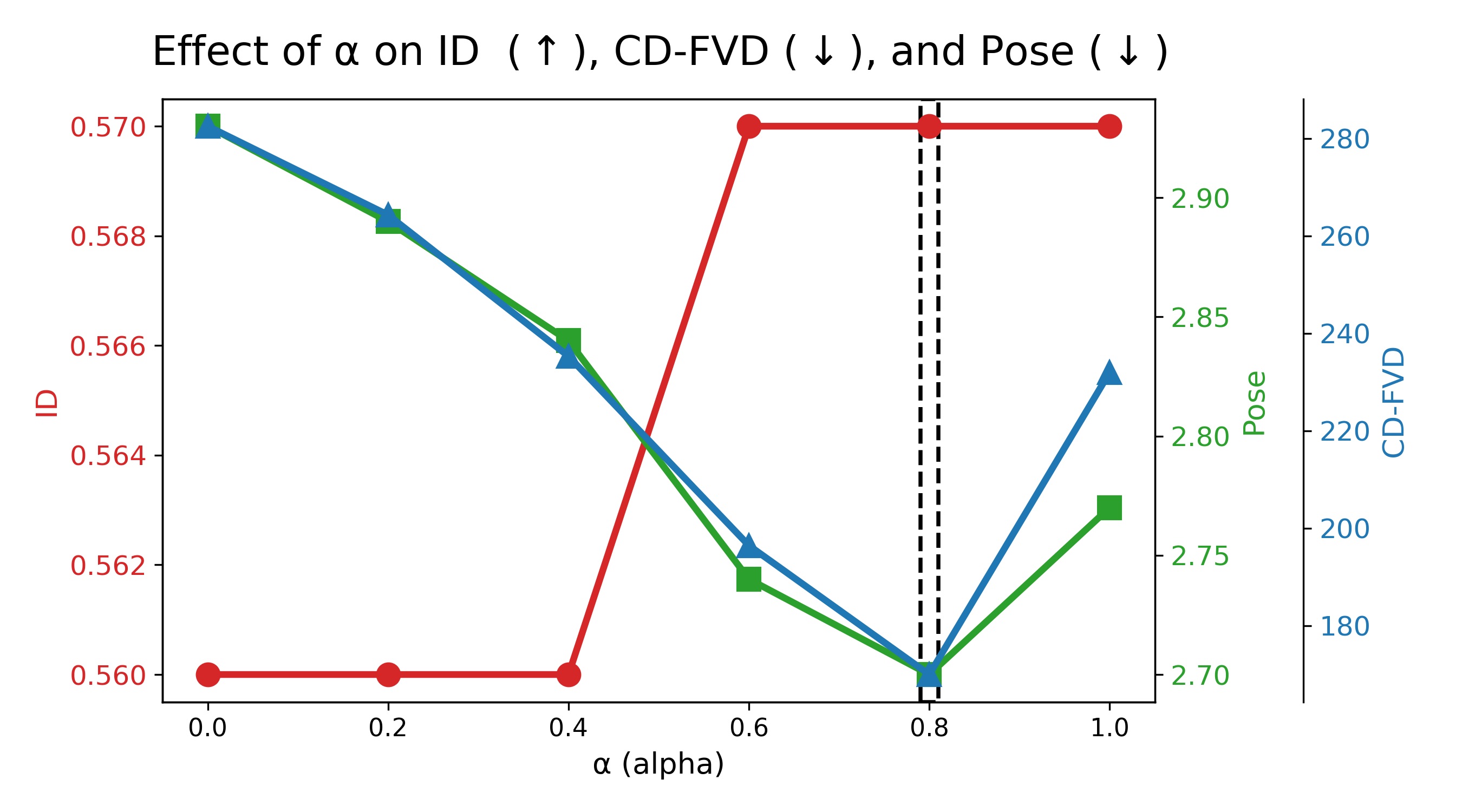}
        \caption{Effect of $\alpha$ (blending ratio).}
        \label{fig:alpha}
    \end{subfigure}
    \caption{Parameter ablations on frequency split ratio $\rho$ and blending ratio $\alpha$.}
    \label{fig:rho_alpha}
\end{figure}

We conduct hyper parameter analysis on $\rho$ the FSAI split ratio, $\alpha$ the FATS blending ratio which is shown in Fig.~\ref{fig:rho} and Fig.~\ref{fig:alpha} respectively. We find that   values around 0.8 for both parameters yield the highest ID while maintaining considerably lower CD-FVD and Pose errors. Further, we observe that the FVD values for FATS time steps $T_1=10$ and $T_1=20$ are 230 and 639, respectively. These results indicate that, while FATS can improve video fidelity performance, excessive usage can lead to severe degradation. 

\subsection{Ablations on Image Datasets and Hyperparameters}
We provide additional ablations on \textbf{FFHQ} and \textbf{CelebA} to isolate the effects of each module. 
As shown in Table~\ref{tab:celeba_ablation_new} and Table~\ref{tab:ffhq_ablation_new}, these results quantify the impact of TSG and FSAI, confirming that both components contribute positively to frame-level fidelity and identity preservation across datasets. 
While TSG improves pose accuracy, it slightly degrades ID transferability. FSAI compensates for this degradation, restoring ID performance to baseline levels while preserving the pose improvements introduced by TSG.

\begin{table}[!htbp]
\centering
\setlength{\tabcolsep}{2pt} 
\begin{tabular}{lccccc}
\toprule
\multirow{2}{*}{\textbf{Method}} & \multirow{2}{*}{\textbf{FID} $\downarrow$} & \multicolumn{2}{c}{\textbf{ID retrieval $\uparrow$}} & \multirow{2}{*}{\textbf{Pose} $\downarrow$} & \multirow{2}{*}{\textbf{Expr.} $\downarrow$} \\
 & & Top-1 & Top-5 & & \\
\midrule
REFace \cite{baliah2025reface} & 5.53 & 95.4\% & 98.7\% & 3.74 & 1.04 \\
\cite{baliah2025reface} + TSG & 4.69 & 94.1\% & 98.2\% & 2.82 & 0.86\\
\cite{baliah2025reface} + TSG + FSAI & 4.29 & 96.2\% & 98.4\% & 2.90 & 0.86 \\
\bottomrule
\end{tabular}
\caption{Ablation study on the FFHQ dataset.}
\label{tab:ffhq_ablation_new}
\end{table}

\begin{table}[!htbp]
\centering
\setlength{\tabcolsep}{2pt} 
\begin{tabular}{lccccc}
\toprule
\multirow{2}{*}{\textbf{Method}} & \multirow{2}{*}{\textbf{FID} $\downarrow$} & \multicolumn{2}{c}{\textbf{ID retrieval $\uparrow$}} & \multirow{2}{*}{\textbf{Pose} $\downarrow$} & \multirow{2}{*}{\textbf{Expr.} $\downarrow$} \\
 & & Top-1 & Top-5 & & \\
\midrule
REFace\cite{baliah2025reface} & 6.09 & 98.8\% & 99.6\% & 3.51 & 0.96\\
\cite{baliah2025reface} + TSG & 5.48 & 94.8\% & 97.8\% & 2.52 & 0.78\\
\cite{baliah2025reface} + TSG + FSAI & 5.42 & 96.2\% & 98.4 \%& 2.63 & 0.77 \\
\bottomrule
\end{tabular}
\caption{Ablation study on the CelebA dataset.}
\label{tab:celeba_ablation_new}
\end{table}

\subsection{Limitation}
We observe some flickering remains in our VFace outcomes. However, our method substantially improves temporal consistency over the base image-based models. Failure cases, such as severe occlusions, large appearance gaps, and identity leakage, mainly arise from limitations of the underlying models (see Fig.~\ref{fig:failure_case}).

\begin{figure*}[!htb]
  \centering
  \animategraphics[width=0.9\linewidth,loop,poster=first]{10}{Videos/Failure_case/frame_0000}{01}{24}
  \caption{Some failure cases with occlusion and high pose variations. Play the video with adobe reader.}
  \label{fig:failure_case}
\end{figure*}

\subsection{Ethics and Societal Impact}

Our goal is to make face swapping and generation more accessible and creative for all users. However, we acknowledge the risk of misuse, including the creation of misleading or harmful media. Such techniques are required to be able to create synthetic data responsibly for learning deepfake detectors under varied scenarios. To promote responsible use, we encourage the development of safeguards such as bias detection, misuse prevention, and transparency mechanisms.

\subsection{Adaptability to Other Image Based Face Swapping Methods}
\begin{table}[!htp]
\centering
\setlength{\tabcolsep}{1pt}
\scalebox{0.8}{
\begin{tabular}{lccccccc}
\toprule
\multirow{2}{*}{\textbf{Method}}  & \multirow{2}{*}{\textbf{CD-FVD}$\downarrow$} & \multirow{2}{*}{\textbf{FVD}$\downarrow$}    &   \multirow{2}{*}{\textbf{ID Sim.}$\uparrow$}  & \multicolumn{2}{c}{\textbf{ID retrieval $\uparrow$}}   & \multirow{2}{*}{\textbf{Pose}$\downarrow$}   &  \multirow{2}{*}{\textbf{Expr.}$\downarrow$}    \\ 
 &&&&   Top-1 &  Top-5 \\
\midrule
Face Adapter& 426.51  & 435.35 & 0.43 & 92.48\% & 98.33\% & 4.58 & 1.38 \\
\text{[14]}+VFace & 358.34 & 358.77 & 0.43 & 91.67\% & 98.12\% & 4.30 &  1.34 \\
\bottomrule
\end{tabular}}
\caption{VFace Implementation on FaceAdapter.}
\vspace{-1em}
\label{tab:adaptability}
\end{table}

We include results on \textbf{FaceAdapter}\cite{han2024faceadapterpretraineddiffusion}, a recent diffusion-based image face swapping approach, which benefits from our modules, demonstrating the broader applicability of our method. We used the same hyperparameter settings as in ReFace in Table~\ref{tab:adaptability}; however, due to differences in model implementations and underlying diffusion architectures, further hyperparameter tuning may improve performance.


\end{document}